\documentclass[5p]{elsarticle} 

\usepackage{amssymb}

\usepackage{microtype}
\usepackage{lineno}
\usepackage{algorithm}
\usepackage{algorithmic}

\usepackage{bm}
\usepackage{amssymb}
\usepackage{mathrsfs}
\usepackage{appendix}
\usepackage{pifont}
\usepackage{booktabs}
\journal{Nuclear Physics B}

\begin{document}

\begin{frontmatter}



\title{See then Tell: Enhancing Key Information Extraction with Vision Grounding}


%
%

\author[1]{Shuhang Liu\corref{cor1}}
\author[2]{Zhenrong Zhang\corref{cor1}}
\cortext[cor1]{Equal contribution.}

\author[1]{Pengfei Hu}
\author[2]{Jiefeng Ma}
\author[1]{Jun Du}
\author[1]{Qing Wang\corref{cor2}}
\cortext[cor2]{Corresponding author.}

\author[2]{Jianshu Zhang}
\author[2]{Chenyu Liu}

\address[1]{University of Science and Technology
	of China, Hefei, Anhui, China}
\address[2]{iFLYTEK Research, Hefei, Anhui, China}

\begin{abstract}
In the digital era, understanding visually rich documents that combine text, complex layouts, and imagery is crucial. Traditional Key Information Extraction (KIE) approaches  heavily rely on Optical Character Recognition (OCR) tools, making them vulnerable to cascading recognition errors that can severely degrade overall performance. OCR-free models address these issues but often lack vision grounding. Recent methods incorporate explicit coordinate outputs, yet depend on downstream coordinate annotations that are not always available in real-world settings.
In this paper, we introduce STNet  ($\textbf{S}$ee then $\textbf{T}$ell Net), an end-to-end model that jointly produces textual answers and their corresponding vision grounding. Central to STNet is a novel \texttt{<see>} token, prepended to each response, which implicitly encodes the physical coordinates. During generation, \texttt{<see>} directs the model first to $\mathit{see}$ — attending to image regions relevant to the question  — and then to $\mathit{tell}$, emitting the textual answer.
To enhance the model's $\mathit{see}$ capabilities, we collect extensive structured table recognition datasets and leverage GPT-4 to develop TVG ($\textbf{T}$ableQA with $\textbf{V}$ision $\textbf{G}$rounding), a dataset of Question Answering (QA) pairs annotated with vision grounding.
Our approach achieves state-of-the-art performance under comparable backbone on public KIE benchmarks including CORD, SROIE, and DocVQA, and generalizes well without access to downstream coordinate annotations during fine-tuning. Moreover, the proposed vision grounding mechanism can be integrated into Multimodal Large Language Models (MLLMs) like Qwen2-VL, improving zero-shot KIE. The code and dataset will be made publicly available.
\end{abstract}


\end{frontmatter}

\begin{figure}[t]
	\centering
	\includegraphics[width=1\linewidth]{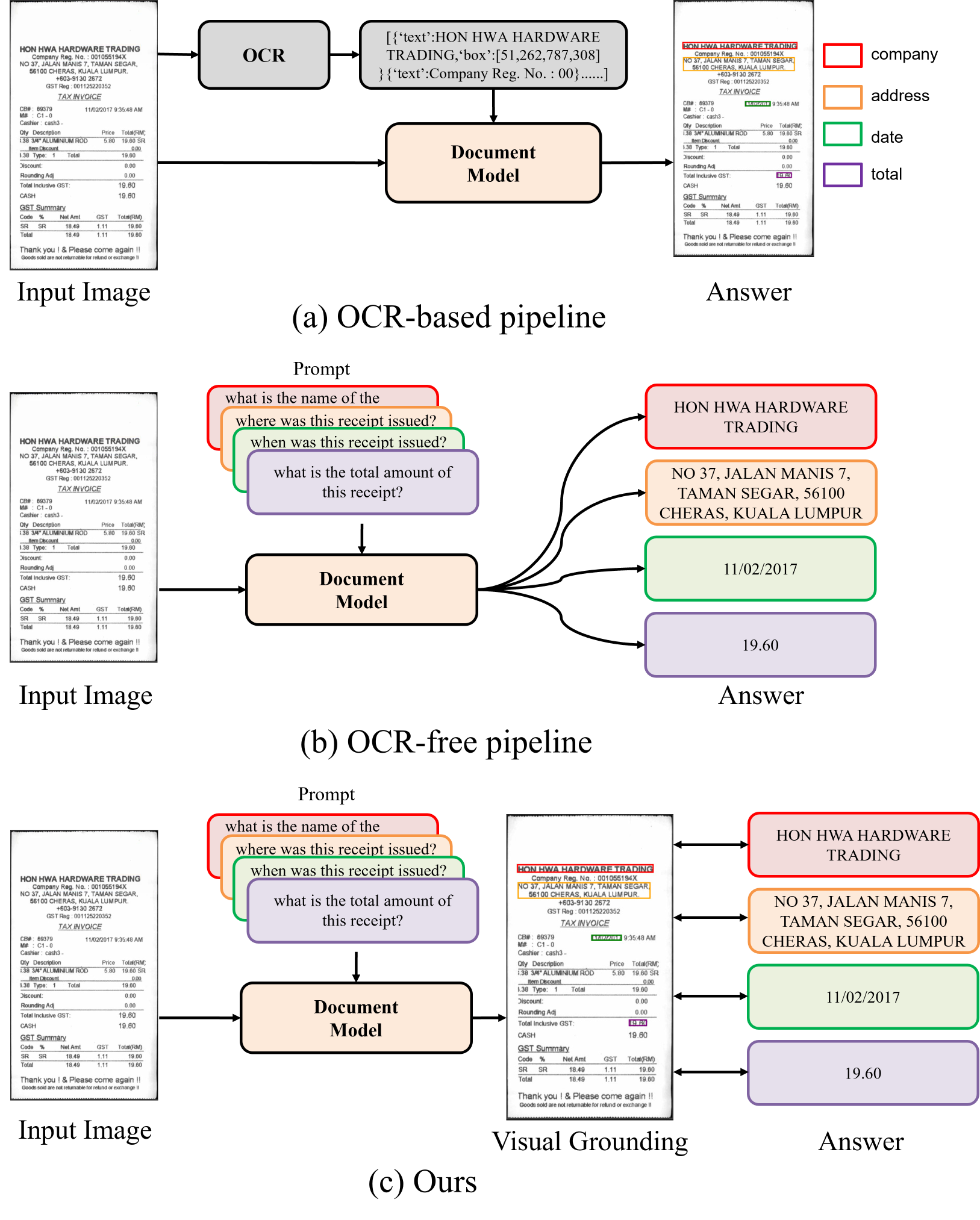}
	\caption{The illustration of contemporary KIE pipelines.}
	\label{fig:pipelines}
\end{figure}
\section{Introduction}

Visually rich document is a type of medium that centers around text while also incorporating the layout and related visual imagery. In the digital information age, many documents are digitized and saved as images.  Understanding document images plays a pivotal role across multiple domains such as document analysis~\cite{2021-CoRR-DocumentAI}, document retrieval~\cite{2020-EMNLP-DocumentRetrieval}, and office robotic process automation~\cite{2020-ACIT-robotic}. This comprehension significantly bolsters the efficiency and accuracy of information processing.
Key Information Extraction (KIE) aims to locate, analyze, and extract key entities (like names, dates, and numerical data) from documents. KIE has become a key part of the document understanding field.

As illustrated in Figure~\ref{fig:pipelines}, existing methods for KIE can be broadly categorized into two groups. Traditional methods~\cite{2022-AAAI-BROS,2022-ACM-LayoutLMv3,2023-TMM-graphdoc,2021-ICCV-DocFormer,Neurocomputing2023graphrevisedie, Neurocomputing2024multimodal,Neurocomputing2025crmsp,Neurocomputing2025seg} rely on Optical Character Recognition (OCR) engines to first extract text and coordinate information from document images, which are then fed into a document model for analysis and classification. However, this pipeline is heavily dependent on the OCR engine, resulting in additional latency and computational costs. Furthermore, errors originating from the OCR step can propagate to the document model, thereby deteriorating overall performance. Recent advancements in document understanding~\cite{2022-ECCV-Donut,2023-ICCV-SeRum} have introduced end-to-end image-to-text paradigms. These methods enable document models to directly process document images without explicit OCR. To achieve this, they leverage the Transformer architecture~\cite{transformer} to decode OCR results during the pre-training stage, endowing the document model with reading capabilities. These OCR-free approaches show strong performance across various document understanding tasks. Nevertheless, the KIE task differs from typical Visual Question Answering (VQA) tasks~\cite{2019-CVPR-TextVQA,ScienceQA} due to the strong correspondence required between the extracted information and the visual content of the document. While recent methods~\cite{2024omniparser,2025omniparserv2} attempt to output explicit coordinates, they often rely on downstream coordinate annotations that are costly and not always available in real-world settings.
These limitations motivate the design of a module that aligns predicted textual answers with the vision grounding, ideally without requiring downstream coordinate annotations.

Currently, most document KIE datasets, such as Doc-VQA~\cite{docvqa} and WikiTableQuestions~\cite{WikiTableQuestions},  purely offer simple plain text Question Answering (QA) pairs without vision grounding for each answer within the image context. With the rise of Large Language Models (LLMs) like ChatGPT~\cite{2020-NeurIPS-ChatGPT} and GPT-4~\cite{2023-CoRR-gpt4}, recent work~\cite{2023-CoRR-ChartLlama,2023-CoRR-TableLlama} has explored using LLMs to generate domain-specific instruction tuning data. Inspired by this, we present an automated processing pipeline that leverages GPT-4 to generate KIE data with robust vision grounding for the document domain. This is expected to enhance KIE performance by providing precise and dependable vision grounding.

In this work, we introduce a novel end-to-end model named STNet (\textbf{S}ee then \textbf{T}ell Net), which can simultaneously provide answers and corresponding vision grounding. We design a \texttt{<see>} token, together with a dedicated physical decoder, to implicitly encode and interpret the coordinates of relevant regions within the image. In downstream tasks, we simply place the \texttt{<see>} token at the beginning of the answer text, thereby providing vision grounding for the answer. To enhance the model's $\mathit{see}$ capabilities, we collect a substantial number of highly structured table recognition datasets, such as PubTables1M~\cite{pubtables} and iFLYTAB~\cite{2024-PR-SEMv2}. Leveraging the powerful text understanding capabilities of GPT-4, we construct a TVG (\textbf{T}ableQA with \textbf{V}ision \textbf{G}rounding) dataset. This dataset not only provides the related plain text QA pairs but also includes the specific vision grounding of the QA pairs within the image.
We validate STNet on publicly available datasets such as CORD~\cite{cord}, SROIE~\cite{SROIE}, and DocVQA~\cite{docvqa}, achieving state-of-the-art results under comparable backbone. Notably, STNet generalizes well without access to downstream coordinate annotations during fine-tuning. Furthermore, the proposed vision grounding mechanism can be seamlessly integrated into Multimodal Large Language Models (MLLMs), such as Qwen2-VL, leading to improved zero-shot KIE performance. The main contributions of this paper are as follows:

\begin{itemize}
	\item We introduce STNet, a novel end-to-end model that not only provides textual answers but also excels in offering vision grounding, enabled by our specially designed \texttt{<see>} token and specialized physical decoder.
	
	\item We introduce a GPT-4-driven automated QA pair generation method, creating the TVG dataset. This dataset comprises QA pairs with precise vision grounding, essential for enhancing visual comprehension.
	
	\item Experimental results on publicly available datasets such as CORD, SROIE, and DocVQA demonstrate that our method achieves state-of-the-art performance and generalizes well without requiring downstream coordinate annotations during fine-tuning.
	
	\item We demonstrate that the proposed vision grounding mechanism can be integrated into MLLMs such as Qwen2-VL, yielding consistent improvements on zero-shot KIE. 
	
\end{itemize}

\section{Related Work}

Early KIE algorithms used rule-based methods, heavily relying on prior knowledge. These approaches were limited to fixed-format documents and lacked robustness for diverse real-world applications. The rapid development of deep learning has brought superior solutions to document understanding, which can be primarily categorized into OCR-based and OCR-free methods.

\subsection{OCR-based Methods}
OCR-based methods require an OCR engine to extract text and coordinate information from visually rich document images as input for document models. The LayoutLM family~\cite{2020-ACM-LayoutLM} introduces a pre-training framework that combines text and layout features, with LayoutLMv2~\cite{2021-IJCNLP-LayoutLMv2} enhancing representation capabilities through spatial-aware self-attention and tasks like text-image alignment and text-image matching. LayoutLMv3~\cite{2022-ACM-LayoutLMv3} further advances this approach by reducing visual feature extraction costs with patch encoding and introducing masked image modeling and word-patch alignment tasks.
DocFormer~\cite{2021-ICCV-DocFormer} integrates visual and spatial information into each Transformer layer using a self-attention encoder. GraphDoc~\cite{2023-TMM-graphdoc} employs BERT and Swin Transformer for semantic and visual encoding, respectively, with an attention-based graph network for localized feature interaction. 
Despite their commendable performance, these models exhibit a strong reliance on OCR tools and subsequent post-processing, which often introduces non-negligible latency~\cite{2022-ECCV-Donut}. Moreover, the dependence on OCR makes them vulnerable to cascading recognition errors that can severely degrade overall performance in real-world settings.

\begin{figure*}[t]
	\centering
	\includegraphics[width=\textwidth]{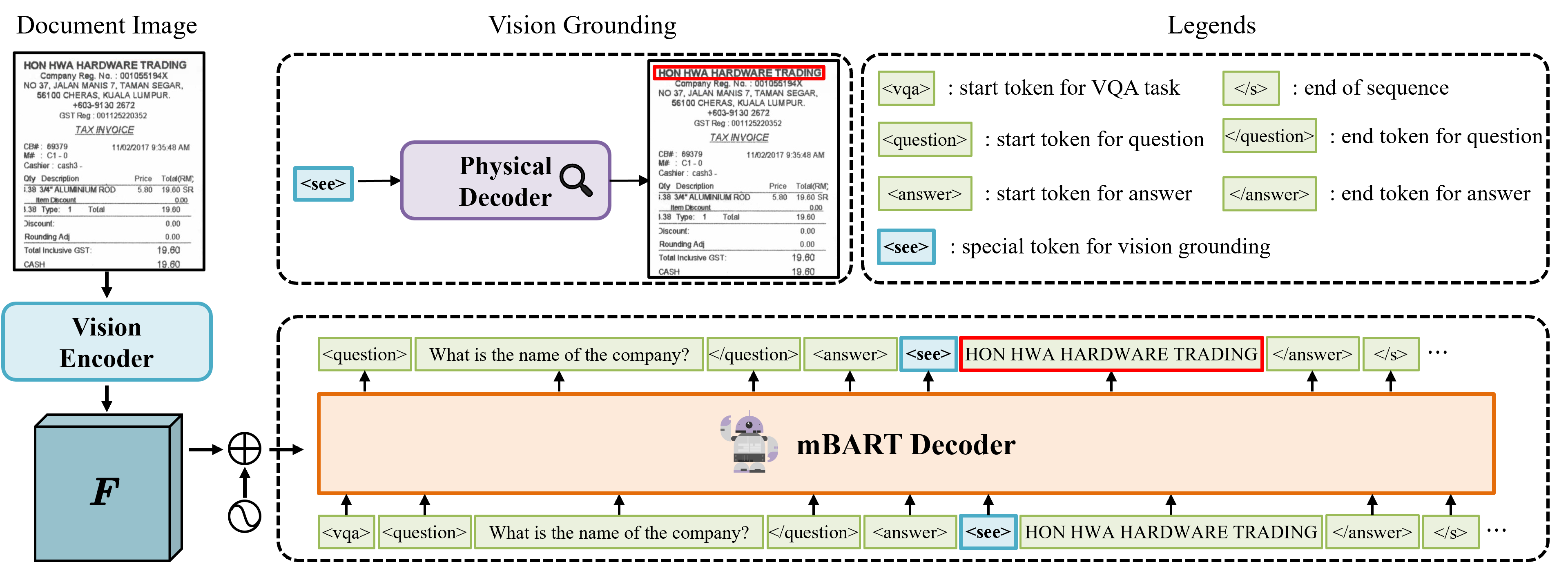}
	\caption{The overall architecture of STNet. It mainly consists of a vision encoder and a text decoder. Our text decoder's output answer sequence includes a special \texttt{<see>} token. The physical decoder is designed to decode the hidden states corresponding to the \texttt{<see>} token to obtain coordinates for vision grounding.}
	\label{fig:STNet}
\end{figure*}
\subsection{OCR-free Methods}
OCR-free methods aim to remove reliance on OCR modules, enabling faster inference with fewer parameters. For example, Donut~\cite{2022-ECCV-Donut} uses the Swin Transformer to encode image patches and BART-like Transformers to generate text sequences, introducing a prompt mechanism to switch between tasks. This end-to-end approach simplifies the model architecture and achieves cost-effectiveness by directly mapping input images into structured outputs. Pix2Struct~\cite{2023-ICML-Pix2Struct} extends these improvements by scaling up pre-training data and tasks, while SeRum~\cite{2023-ICCV-SeRum} employs selective region concentration to enhance precision and speed. These methods streamline information processing and accelerate reasoning, making them highly effective in KIE.
Recently, some researchers have also attempted to provide vision grounding for answers, further improving the effectiveness of these models. 
CREPE~\cite{2024-CoRR-CREPE} employs a multi-head architecture, where a specialized head is designed to predict text coordinates after the answer text. However, using generated text to assist in location inference (``tell then see") provides limited improvement for KIE itself, as will be validated through comparisons with our ``see then tell" approach.
OmniParser~\cite{2024omniparser} employs a two-stage decoding scheme that generates structured points as intermediate adapters to jointly output text and coordinates for KIE, while OmniParser V2~\cite{2025omniparserv2} further improves this with Structured-Points-of-Thought prompting and a token-router-based shared decoder, achieving better efficiency and accuracy. However, the reliance on intermediate structured-point generation introduces redundancy, error accumulation, and inference latency, and the explicit coordinate outputs require downstream coordinate annotations during fine-tuning, which are not always available in real-world scenarios.

\subsection{Leveraging LLMs for Dataset Construction}
High-quality datasets are crucial for improving model performance, yet existing ones often fail to meet emerging needs, and manual annotation is costly.  With the advent of LLMs like ChatGPT~\cite{2020-NeurIPS-ChatGPT} and GPT-4~\cite{2023-CoRR-gpt4}, researchers have begun leveraging their robust linguistic and coding capabilities to efficiently process large volumes of data and construct new datasets. WizardLM~\cite{2023-CoRR-WizardLM} and GPT-4-LLM~\cite{2023-CoRR-Instruction_gpt4} have validated the effectiveness of using ChatGPT and GPT-4 to generate instruction fine-tuning datasets. ChartLlama~\cite{2023-CoRR-ChartLlama} utilizes GPT-4 to generate chart images along with diverse and precise QA pairs, thereby aiding in the training of models for comprehensive chart understanding. TableLlama~\cite{2023-CoRR-TableLlama} has achieved similar advancements in the domain of table data. However, the WizardLM and GPT-4-LLM datasets are restricted to textual data only.
While TableLlama and ChartLlama are multimodal, they lack spatial annotations for the QA pairs, limiting their utility for vision grounding. To address these limitations, we propose a novel automated QA pair generation method to construct the TVG dataset, which includes QA pairs along with their corresponding spatial information.

\section{Task Definition}
Given a document image $\bm{I} $ and a question sequence \( \bm{Q} = \{ \bm{q}_i \in \mathbb{R}^v \mid i = 1, \dots, T_q \} \) as a prompt, our objective is to enable the model to predict the answer sequence \( \bm{A} = \{ \bm{a}_i \in \mathbb{R}^v \mid i = 1, \dots, T_a \} \) for completing KIE. Here, $T_q$ denotes the length of the question sequence, $T_a$ denotes the length of the answer sequence, and $v$ is the size of the token vocabulary. Previous methods have achieved remarkable results using this format.
In contrast, we divide this process into two distinct phases: $\mathit{see}$ and $\mathit{tell}$. In the $\mathit{see}$ phase, we output a \texttt{<see>} token that implicitly encodes the physical coordinates $\bm{p} = \{p_j \in \mathbb{N} \mid j = 1, \dots, 8\}$, which represent a four-point polygon, defining the physical location in the document image associated with $\bm{Q}$. This differs from logical locations, such as row and column coordinates. We choose four-point polygons over bounding boxes, as polygons more effectively handle warped or rotated text. Subsequently, in the $\mathit{tell}$ phase, the model generates the answer text following the \texttt{<see>} token.

\section{Methodology}
As illustrated in Figure~\ref{fig:STNet}, STNet is built on Donut ~\cite{2022-ECCV-Donut} and consists of two primary modules: a vision encoder and a text decoder. The vision encoder is tasked with processing image features which are subsequently interpreted by the text decoder to formulate the answer sequence $\bm{A}$. Our model introduces a novel \texttt{<see>} token that implicitly encodes physical coordinates at the beginning of $\bm{A}$. Specifically, we design a dedicated physical decoder to extract these coordinates and propose a corresponding $\mathit{see}$ loss to supervise this encoding. The \texttt{<see>} token with physical location information effectively guides the subsequent output of the answer text, ensuring ``see then tell''. Due to the lack of suitable datasets to train \texttt{<see>}, we accumulate a large amount of structured table recognition data and construct the TVG dataset utilizing GPT-4. More details are elaborated in subsequent sections.
\subsection{Vision Encoder}
The vision encoder transforms the input document image  $\bm{I}$ into a feature map $ \bm{F} \in \mathbb{R}^{H \times W \times D} $. This feature map is subsequently serialized into a set of embeddings $ \bm{Z} = \{ \bm{z}_i \in \mathbb{R}^D \mid i = 1, \dots , N \} $, where $ N $ represents the size of the feature map and $ D $ is the dimension of the encoder's latent vectors.
Following Donut, we adopt the Swin Transformer~\cite{2021-ICCV-swintransformer} as our primary vision backbone due to its superior performance demonstrated in previous studies. 
Additionally, we incorporate positional encoding~\cite{transformer} into $\bm{F}$ to produce the final vision embeddings $\bm{Z}$, enhancing the model’s perception of location.

\subsection{Text Decoder}
Similar to Donut, we utilize the BART~\cite{2020-ACL-BART} decoder to generate the answer sequence $\bm{A}$, which is conditioned on the $\bm{Z}$ and prompted by the question sequence $\bm{Q}$. Since STNet is trained to predict the next token  like LLMs~\cite{2023-CoRR-gpt4}, the training objective is to minimize the negative log-likelihood of the target sequence.

\begin{equation}
	\mathscr{L}_{\mathrm{lm}}= -\frac{1}{T_a} \sum_{i=1}^{T_a} \log P\left(\bm{a}_i \mid \bm{Z}, \bm{Q}, \bm{a}_{1: i}\right)
\end{equation}

\subsection{Physical Decoder}

We explore the fundamental human cognitive process of ``see then tell'', where individuals first $\mathit{see}$ — gathering visual information and contextual insights — and then $\mathit{tell}$ — constructing responses. This sequence notably enhances the accuracy and relevance of interactions. To effectively mirror this intuitive cognitive pattern, our proposed method adopts a two-phase output strategy. The \texttt{<see>} token initiates the $\mathit{see}$ phase to perceive location information related to $\bm{Q}$ in the document image, followed by the $\mathit{tell}$ phase that outputs the answer text. We design a physical decoder that decodes the hidden states \( \bm{H} = \{\bm{h}_i \in \mathbb{R}^D \mid i = 1, \dots, T\} \) extracted from the final layer of our text decoder, specifically corresponding to the \texttt{<see>} token, allowing us to obtain the polygon coordinates $\bm{p}$ within the image context for vision grounding. To facilitate this prediction, we employ a quantization strategy utilizing a specialized vocabulary composed of 1,000 unique tokens, ranging from \texttt{<0>} to \texttt{<999>}, collectively denoted as \( \bm{Loc} \in \mathbb{R}^{1000 \times D} \).
For each coordinate  $p_j$  within a polygon \( \bm{p}_i \), its associated hidden state \( \bm{h}_i \) undergoes a linear transformation, producing  $\bm{h}_i^{p_j}$ as a query against
the vocabulary \( \bm{Loc} \). 
The final determination of \( p_j \)'s position is computed based on the expected location derived from the probability distribution over \( \bm{Loc} \), as provided by $\bm{h}_i^{p_j}$, divergent from previous direct classification methods~\cite{2022-ICLR-pix2seq} over a location vocabulary:

\begin{equation}
	\bm{h}_i^{p_j}=\mathrm{Linear} \left(\bm{h}_i\right)  
\end{equation}
\begin{equation}
	\bm{b}^{p_j}=\mathrm{softmax} \left(\bm{h}_i^{p_j} \bm{L o c}^{\top}\right)   
\end{equation}
\begin{equation}
	E\left(p_j\right)=\sum_{i=0}^{999} i \cdot b_i^{p_j}    
\end{equation}

Here, $\bm{b}^{p_j} \in \mathbb{R}^{1000}$ represents the probability distribution for the position of $p_j$. The polygon regression loss of   $\mathit{see}$  is defined as follows:
\begin{equation}
	\mathscr{L}_{\mathrm{see}}=\frac{1}{8} \sum_{j=1}^8\left(E\left(p_j\right)-p_j^*\right)^2
\end{equation}

where $p_j^*$ represents the ground truth label.

\subsection{TVG Dataset}

\begin{figure*}[t]
	\centering
	\includegraphics[width=0.8\textwidth]{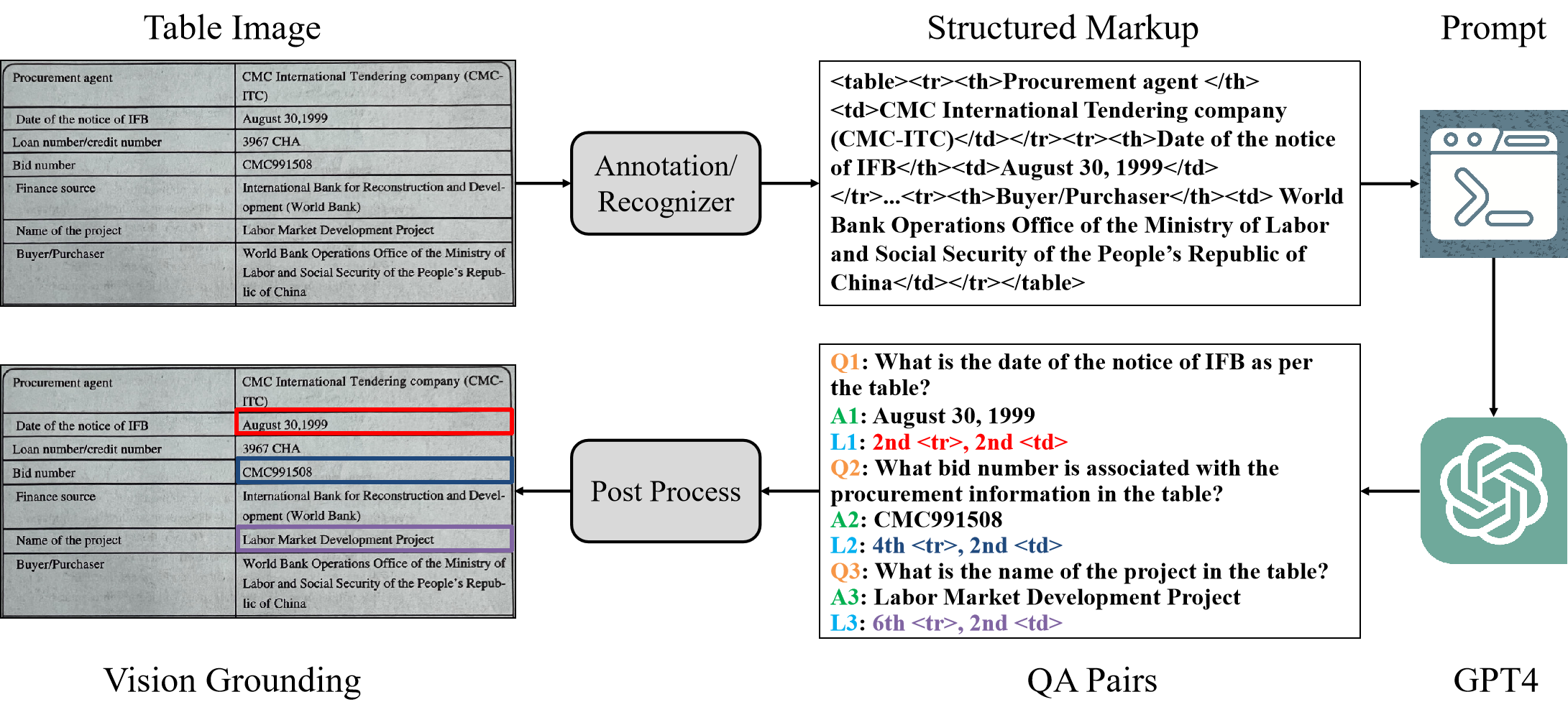}
	\caption{TVG dataset construction pipeline.}
	\label{fig:TVG}
\end{figure*}

Current document datasets, such as DocVQA~\cite{docvqa} and WikiTableQuestion~\cite{WikiTableQuestions}, only offer plain text QA pairs, constraining the enhancement of STNet's ability to $\mathit{see}$.
Inspired by recent studies~\cite{2023-CoRR-WizardLM,2023-CoRR-ChartLlama,2023-CoRR-TableLlama} which leverage LLMs for dataset construction, we propose a comprehensive GPT-4-based method, as depicted in Figure~\ref{fig:TVG}, to automatically construct QA datasets for document images.

\subsubsection{Details of Data Source}
Tables, as a special form of document image, can be described using structured markup such as HTML, and high-quality table data can be readily sourced from online resources. To this end, we have compiled several table recognition datasets, including PubTables1M~\cite{pubtables} and iFLYTAB~\cite{2024-PR-SEMv2}.
\begin{itemize}
\item PubTables1M is a large-scale table recognition dataset sourced from the PubMed Central Open Access (PMCOA) database. This dataset includes detailed annotations for projected row headers and bounding boxes for all rows, columns, and cells, including blank cells. Additionally, it introduces a novel canonicalization procedure aimed at correcting over-segmentation. This procedure ensures that each table is presented with a unique and unambiguous structural interpretation. Through these detailed annotations, we transform the tables into an HTML format.

\item The iFLYTAB dataset comprises 12,104 training samples and 5,187 testing samples. It offers comprehensive annotations for each table image, including both physical coordinates and detailed structural information. This dataset includes not only axis-aligned digital documents but also images captured by cameras, which present more significant challenges than PubTables1M due to their complex backgrounds and non-rigid image deformations. Although it lacks textual annotations, we have addressed this limitation by using PaddleOCR~\cite{ppocr} for text recognition, subsequently converting the tables into HTML format.
\end{itemize}

\subsubsection{Generation Prompt}
\begin{figure*}[h]
	\centering
	\includegraphics[width=\textwidth]{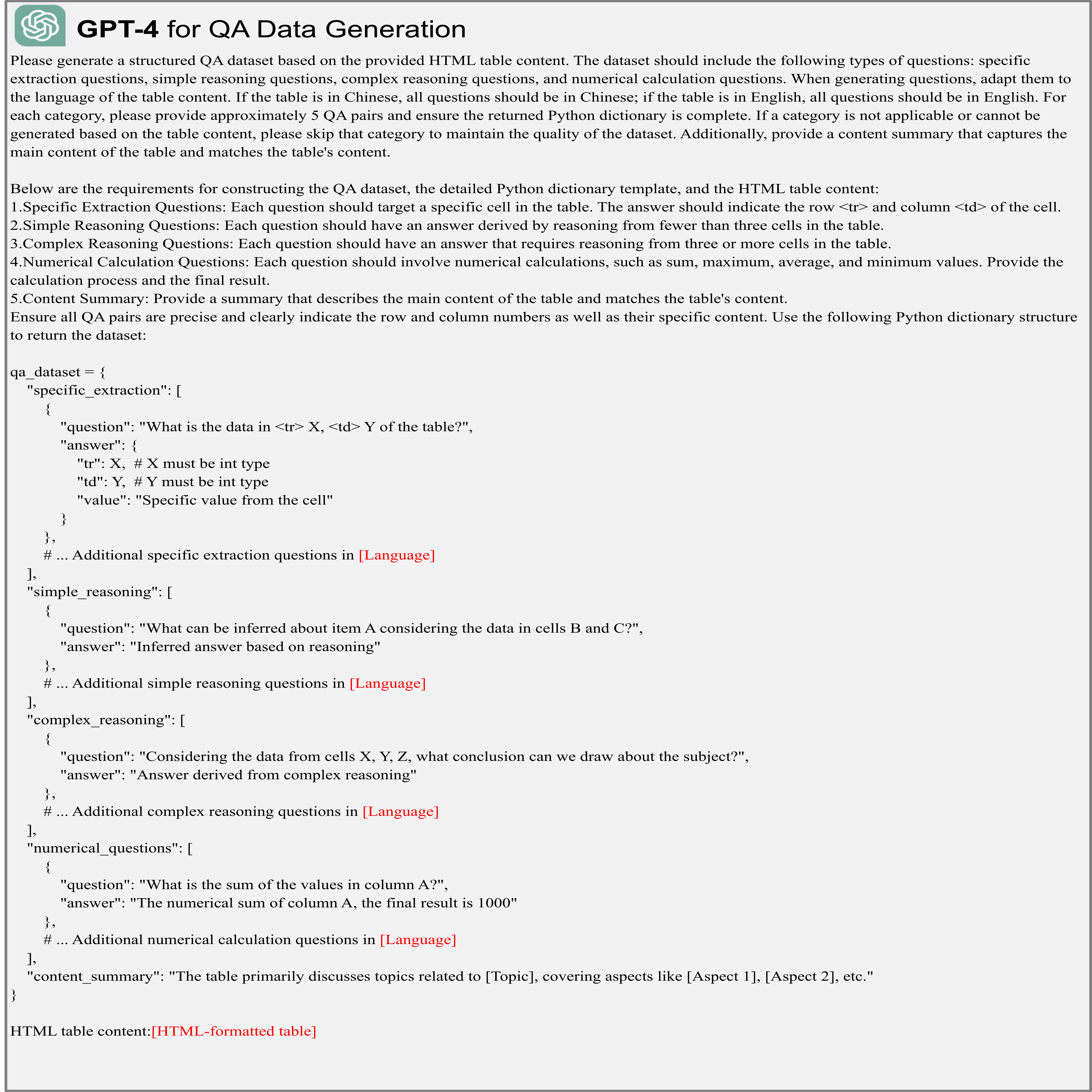}
	\caption{The prompt template for QA data generation.}
	\label{fig:prompt}
\end{figure*}

As shown in Figure~\ref{fig:prompt}, we present a standardized prompt template designed for QA data generation using GPT-4, which requires structured HTML table sequences as input. The text in black represents fixed components of the prompt, and the text within red brackets  requires specific input. For example, $[Language]$ specifies the language in which the QA pairs should be generated.
We instruct GPT-4 to generate five types of questions: specific extraction, simple reasoning, complex reasoning, numerical questions, and content summary.
\begin{itemize}
	\item \textbf{Specific Extraction}: Each specific extraction question should target a specific cell in the table. The answer should indicate the row \texttt{<tr>} and column \texttt{<td>} of the cell.

	\item \textbf{Simple Reasoning}: Each simple reasoning question should have an answer derived by reasoning from fewer than three cells in the table.
	
	\item \textbf{Complex Reasoning}: Each complex reasoning question should have an answer that requires reasoning from three or more cells in the table.
	
	\item \textbf{Numerical Questions}: Each numerical question should involve numerical calculations, such as sum, maximum, average, and minimum values. Provide the calculation process and the final result.
	
	\item \textbf{Content Summary}: Each content summary needs to provide a summary that describes the main content of the table and matches the table's content.
\end{itemize}

\subsection{Post Process}
\begin{figure*}[h]
	\centering
	\includegraphics[width=\linewidth]{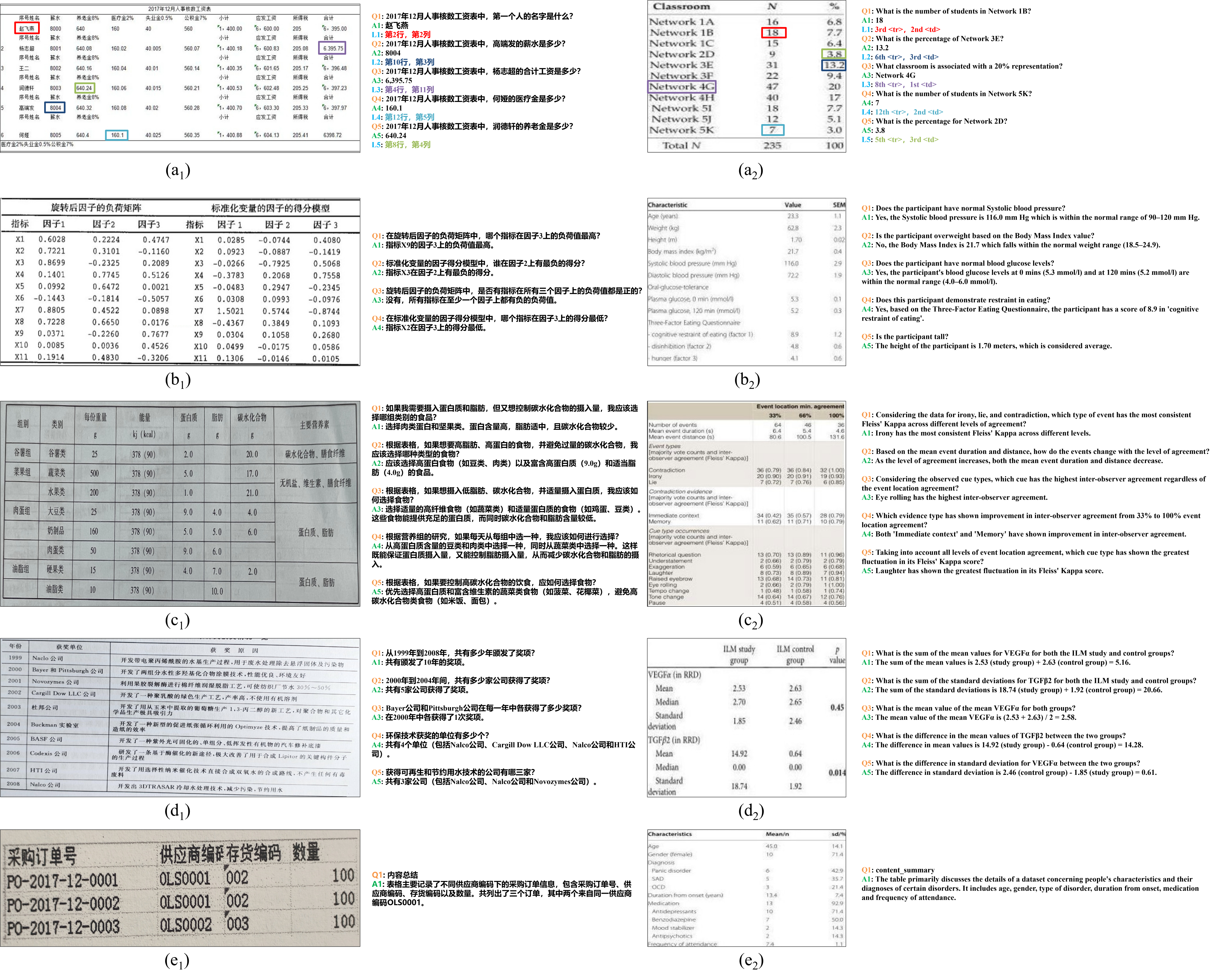}
	\caption{ Some examples of the TVG dataset. a* refer to specific extraction, where L indicates the logical location and its text color corresponds to the coordinate box in the table image. b* refer to simple reasoning, c* refer to complex reasoning, d* refer to numerical questions, and e* refer to content summary.}
	\label{fig:TVG_sample}
\end{figure*}

As described in the aforementioned prompt template, for specific extraction questions, we require GPT-4 to provide not only the specific value from the cell but also the logical location of the cell, indicating its row and column numbers within the table. The detailed annotations in these table recognition datasets enable us to accurately locate the corresponding cell's real information, including its content and polygon box, based on the logical location. We only retain the QA pair when the value provided by GPT-4 matches the content of the located cell, with the cell polygon box serving as the required physical location $\bm{p}$. This process ensures the generation of high-quality QA data $\{\bm{Q}, \bm{A}, \bm{p}\}$. Ultimately, the TVG dataset we construct comprises 958,000 questions derived from 65,000 table images. It includes 244k specific extraction questions, 293k simple reasoning questions, 191k complex reasoning questions, 166k numerical questions, and 64k content summary. Some examples of them are illustrated in Figure~\ref{fig:TVG_sample}. The whole dataset will be made  publicly available. 

\clearpage

\section{Implementation Details}
\begin{figure}[t]
	\centering
	\includegraphics[width=1\linewidth]{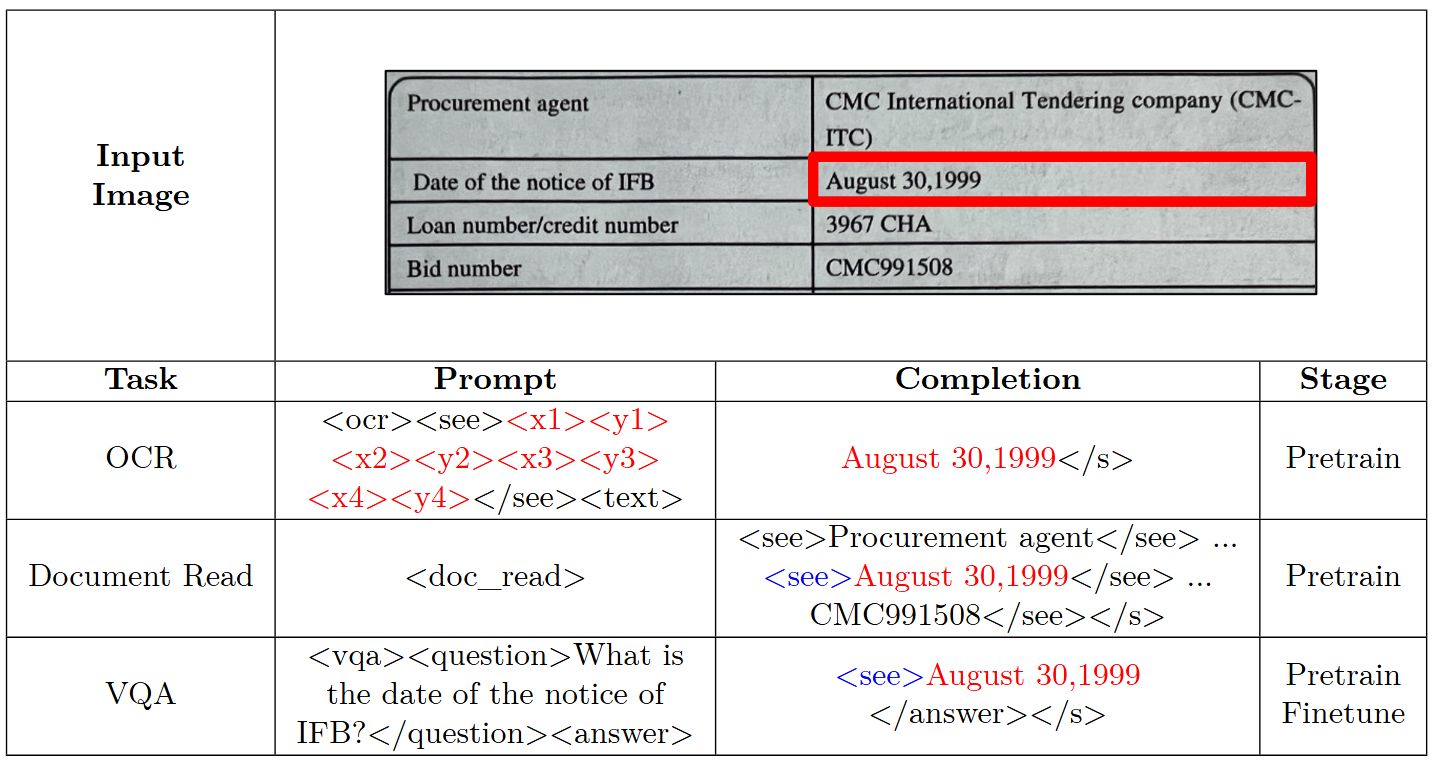}
	\caption{The illustration of the task design.}
	\label{fig:task_design}
\end{figure}
\subsection{Model Architecture}
Our proposed STNet utilizes specific hyperparameters: We set the input image resolution to $1280 \times 960$ and use random padding to maintain the original aspect ratio. The visual backbone's downsampling factor is configured to 32. The feature dimension $D$ is established at 1024. The decoders consist of a stack of 4 identical layers, and the number of multi-heads is set to 16.
\subsection{Pre-training Tasks}

To bolster STNet’s capability to $\mathit{see}$ — we have integrated a multi-task pre-training strategy encompassing three distinct sub-tasks: OCR, Document Read, and VQA, as illustrated in Figure~\ref{fig:task_design}.

\subsubsection{OCR.}
In this task, locations are represented as polygons defined by four coordinate points, with each point discretized into a token ranging from \texttt{<0>} to \texttt{<999>} in $\bm{Loc}$. Given prompts containing such tokens, STNet is trained to generate the corresponding text within the specified regions, effectively emulating the behavior of a traditional OCR engine.

\subsubsection{Document Read.}
To enhance structural understanding, the model is trained to read document content in the conventional reading order. Each text block is preceded by a \texttt{<see>} token, which implicitly encodes its location. The physical coordinates corresponding to \texttt{<see>} are obtained by decoding the hidden states through the specialized physical decoder, thereby reinforcing the alignment between text and its spatial context.

\subsubsection{VQA.}
Expanding beyond conventional VQA tasks, STNet is designed to not only generate a plaintext response but also identify the relevant text coordinates using the \texttt{<see>} token for vision grounding. This approach aligns with the requirements of downstream KIE tasks.

\subsection{Training Strategy}
In the initial pre-training phase, in addition to the previously constructed TVG dataset and its data sources — the training sets from PubTables1M~\cite{pubtables} and iFLYTAB~\cite{2024-PR-SEMv2} — we also employ a synthetic dataset comprising 2.2 million entries in both Chinese and English from SynthDog~\cite{2022-ECCV-Donut}. PubTables1M, iFLYTAB, and SynthDog are used for OCR and document read training, while TVG is utilized for OCR and VQA tasks. 

After pre-training, STNet is fine-tuned on specialized datasets for KIE. Each dataset is tailored to meet the VQA task specifications and is combined with TVG at a 1:1 ratio, with consistent supervision of the $\mathit{see}$ loss on TVG. It ensures that the output \texttt{<see>} token retains its physical location perception ability to guide the output of answer text, even without $\mathit{see}$ supervision on downstream datasets.

We use the Adam optimizer~\cite{2015-ICLR-Adam} with a learning rate of $5\times10^{-5}$. The learning rate is linearly warmed up during the first 10\% of the steps and then linearly decayed. The training is conducted on 4 Tesla V100 48GB GPUs with a total batch size of 28. 
The model is trained for 250 epochs on the SROIE~\cite{SROIE} and CORD~\cite{cord} datasets, and extended to 300 epochs for the DocVQA~\cite{docvqa} dataset.

STNet utilizes two types of loss:  
$\mathscr{L}_{\mathrm{lm}}$ and  $\mathscr{L}_{\mathrm{see}}$. The total loss is computed as a weighted sum of them.
\begin{equation}
	\mathscr{L}_{\mathrm{total}}=\mathscr{L}_{\mathrm{lm}}+\lambda\mathscr{L}_{\mathrm{see}}
\end{equation}
After extensive evaluation, we set  $\lambda =0.001 $.

\subsection{Inference}
During the inference phase, we feed the question sequence $\bm{Q}$ to STNet as a prompt, guiding it to output the answer sequence $\bm{A}$.  Utilizing the hidden states $\bm{H}$ from the text decoder's last layer, we can decode the polygon $\bm{p}$, associated with the \texttt{<see>} token preceding each response in $\bm{A}$. Through the visualization of this vision grounding, we can enhance the interpretability of the answers.

\begin{table*}[t]
	\centering
	\begin{tabular}{lccccccccc}
		\toprule
		Method&Localization  &Downstream& \multicolumn{2}{c}{CORD} && \multicolumn{2}{c}{SROIE} && DocVQA \\
		\cmidrule{4-5} \cmidrule{7-8} \cmidrule{10-10}
		&Ability&Coordinate& F1 & Acc. && F1 & Acc. && ANLS  \\
		
		\midrule
		Donut~\cite{2022-ECCV-Donut} &\ding{56} &\ding{56}& 84.1 & 90.9 && 83.2 & 92.8 && \underline{59.7}  \\
		SeRum~\cite{2023-ICCV-SeRum} &\ding{56} &\ding{56}& 80.5 & 85.8 && 85.8 & 95.4 && -  \\
		CREPE~\cite{2024-CoRR-CREPE} &\ding{52} &\ding{56}& 85.0 & - && - & - && 58.4  \\
		OmniParser~\cite{2024omniparser} &\ding{52} &\ding{52}& 84.8 & 88.0 && 85.6 & 93.6 && -  \\
		 OmniParser V2~\cite{2025omniparserv2} &\ding{52} &\ding{52}& 85.0 & 88.7 && 85.8 & 94.0 && -  \\
		STNet &\ding{52} &\ding{56}& \underline{88.1} & \underline{92.3} && \underline{87.8} & \underline{97.1} && \textbf{63.7} \\
		STNet* &\ding{52} &\ding{52}& \textbf{88.8} & \textbf{93.5} && \textbf{88.3} & \textbf{97.4} && - \\
		\bottomrule
	\end{tabular}
	
	\caption{Comparison with OCR-free SOTA methods across different datasets. Localization Ability indicates whether a method is capable of providing vision grounding of predicted answers. Downstream Coordinate denotes whether the method requires coordinate annotations from downstream datasets during fine-tuning. The field-level F1 scores and tree-edit-distance-based accuracies
		are reported. \textbf{Bold} indicates the best result. \underline{Underline} indicates the second best.}
	\label{tab:results}
\end{table*}
\section{Experiments}
\subsection{Evaluation Benchmarks and Metrics}
To fully demonstrate the effectiveness of STNet, we conduct experiments on three benchmark datasets. 

\subsubsection{SROIE.} 
The SROIE~\cite{SROIE} dataset comprises 973 scanned receipt images, divided into two subsets: 626 images for training and 347 for testing. Each receipt is annotated with four predefined target fields: company, date, address, and total. Additionally, segment-level text bounding boxes and their corresponding transcripts are provided to facilitate the extraction tasks. The primary objective is to accurately map each word to its appropriate field.
For evaluating model performance on the test set, we employ two metrics: the field-level F1 score~\cite{f1} and Tree Edit Distance (TED)-based accuracy~\cite{ted}. To ensure consistency, we adopt the results reported from Donut~\cite{2022-ECCV-Donut} and SeRum~\cite{2023-ICCV-SeRum}.
\subsubsection{CORD.}
The CORD~\cite{cord} dataset serves as a public benchmark comprising 800 training, 100 validation, and 100 testing receipts. The receipts are annotated with 30 types of entities  under 4 categories: menu, void menu, subtotal, and total. A list of text lines with bounding boxes is provided. The evaluation task and metrics for the CORD dataset align with those used for the SROIE dataset.

\subsubsection{DocVQA.}
The DocVQA~\cite{docvqa} dataset comprises 50,000 questions derived from over 12,000 pages across a wide array of documents. The pages are allocated into training, validation, and test sets with an approximate ratio of 8:1:1. Due to the absence of ground truth in the test set, evaluations are performed on the validation set using the ANLS (Average Normalized Levenshtein Similarity).


%
%
%
%
%

\begin{table}[t]
	\centering
	\renewcommand{\arraystretch}{1} 
	\begin{tabular}{lcccc}
		\toprule
		Method&OCR  & CORD && SROIE \\
		\cmidrule{3-3} \cmidrule{5-5}
		&Input& F1 && F1 \\
		\midrule
		BROS~\cite{2022-AAAI-BROS} & Real OCR & 74.7 && -  \\
		
		LayoutLM~\cite{2020-ACM-LayoutLM} & Real OCR & 78.4 && - \\
		
		LayoutLMv2~\cite{2021-IJCNLP-LayoutLMv2} & Real OCR & 78.9 && 61.0  \\
		
		LayoutLMv3~\cite{2022-ACM-LayoutLMv3} & Real OCR & 80.5 && 65.0  \\
		\midrule
		BROS*~\cite{2022-AAAI-BROS} & GT OCR & 96.5 && 96.3  \\
		LayoutLM*~\cite{2020-ACM-LayoutLM} & GT OCR & - && 94.0 \\
		LayoutLMv2*~\cite{2021-IJCNLP-LayoutLMv2} & GT OCR & 95.0 && 96.3  \\
		LayoutLMv3*~\cite{2022-ACM-LayoutLMv3} & GT OCR & 96.6 && -  \\
		
		\midrule
		STNet &No OCR & 88.1 && 87.8  \\
		\bottomrule
	\end{tabular}
	
	\caption{Comparison with OCR-based SOTA methods across different datasets. The field-level F1 scores are reported. Models marked with * utilize the ground-truth text strings and coordinates as inputs \textbf{during evaluation}, 
		while the unmarked counterparts rely on the outputs of real OCR engines. }
	\label{tab:results_OCR}
\end{table}

\subsection{Results}

We compare our approach with various prior methods. To simulate real-world scenarios where downstream datasets such as DocVQA lack coordinate annotations, our STNet is fine-tuned with $\mathit{see}$ supervision applied only to the auxiliary TVG dataset within the mixed training. In contrast, STNet* denotes the variant where $\mathit{see}$ supervision is additionally applied to the downstream datasets. 

\subsubsection{Comparison with OCR-free Methods.}

\begin{figure*}[t]
	\centering
	\includegraphics[width=0.95\textwidth]{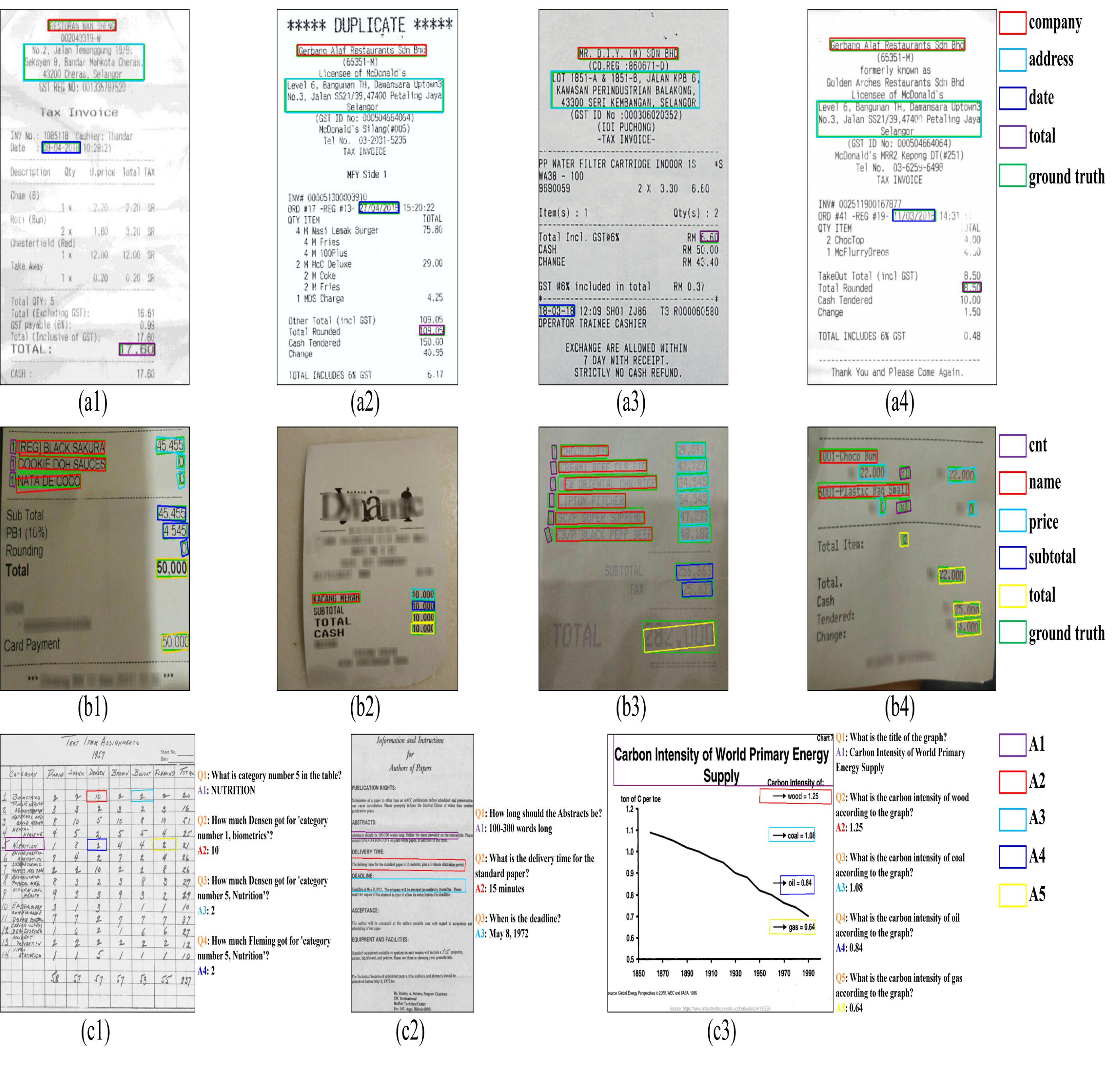}
	\caption{The results of acquiring text coordinates. a* refer to SROIE from STNet*, b* refer to CORD from STNet*, and c* refer to DocVQA from STNet. For SROIE and CORD, different colors of the polygon boxes represent various categories. Each dataset has a different color bar, but both use green to indicate the ground truth. For DocVQA, different colors of the polygon boxes represent answers to different questions. }
	\label{fig:results}
\end{figure*}

As shown in Table~\ref{tab:results}, our approach achieves state-of-the-art performance among OCR-free methods. Here, Localization Ability indicates whether a method is capable of providing vision grounding of predicted answers, while Downstream Coordinate denotes whether the method requires coordinate annotations from downstream datasets during fine-tuning. Notably, the Omniparser series cannot be fine-tuned on datasets such as DocVQA that lack answer coordinate annotations, whereas our STNet remains effective. Although STNet performs slightly below STNet*, the marginal gap highlights its strong generalization ability in scenarios where downstream coordinate annotations are unavailable.
Figure~\ref{fig:results} showcases the outcomes of text coordinates acquisition by our model.

\subsubsection{Comparison with OCR-based Methods.}

OCR-based methods, such as the LayoutLM family~\cite{2020-ACM-LayoutLM,2021-IJCNLP-LayoutLMv2,2022-ACM-LayoutLMv3}, typically use the ground truth of text strings and coordinates during evaluation, as highlighted in their respective papers. This approach simplifies the task to a token classification problem based purely on textual content. To ensure a fair comparison, Donut~\cite{2022-ECCV-Donut} and SeRum~\cite{2023-ICCV-SeRum} re-evaluated these models using state-of-the-art publicly available OCR engines to extract text and corresponding bounding boxes. Following this standard practice, we adopt the results reported in Donut and SeRum and re-evaluate LayoutLMv3~\cite{2022-ACM-LayoutLMv3} under the same conditions.

As shown in Table~\ref{tab:results_OCR}, we compare the performance of these models under two evaluation settings. Notably, models marked with * utilize the ground truth of text strings and coordinates as inputs during evaluation.

It can be observed that these OCR-based methods can achieve satisfactory results when the OCR outputs are entirely accurate. However, producing such precise annotations is costly, and the cascading effects of OCR errors significantly impact the model's performance. In scenarios where only text and bounding boxes extracted by real OCR engines are provided, our STNet demonstrates superior performance compared to these methods.

\subsection{Generalization to Advanced MLLMs}
To assess the generalizability of our approach, we apply it to advanced MLLMs such as Qwen2-VL~\cite{Qwen2-VL}. We integrate our vision grounding module into the MLLM architecture and augment the prompt with the \texttt{<see>} token. The model is trained on the TVG dataset with explicit $\mathit{see}$ supervision on \texttt{<see>} to acquire $\mathit{see}$ capabilities. We adopt the Adam optimizer~\cite{2015-ICLR-Adam} with a learning rate of $1\times10^{-5}$, which is linearly warmed up over the first 5\% of training steps and then decayed linearly. Training is performed on 4 Tesla V100 GPUs (48GB each) with a total batch size of 128 for 3 epochs. To support full-parameter fine-tuning under memory constraints, we employ ZeRO optimization~\cite{zero}.

During evaluation, we follow the zero-shot prompt setting adopted in prior MLLM-based KIE studies~\cite{layoutllm}.
For SROIE and DocVQA, we adopt the same prompt format as used in STNet, with the \texttt{<see>} token appended at the end of the prompt. For key-value annotations in CORD, we format the queries as: “Q: What is the ‘key’? \texttt{<see>} A: ‘value’”. To ensure a fair comparison, we follow previous work by filtering out samples where a single entity corresponds to multiple values.
All evaluations are performed using the ANLS metric. 

\begin{figure*}[ht]
	\centering
	\includegraphics[width=\textwidth]{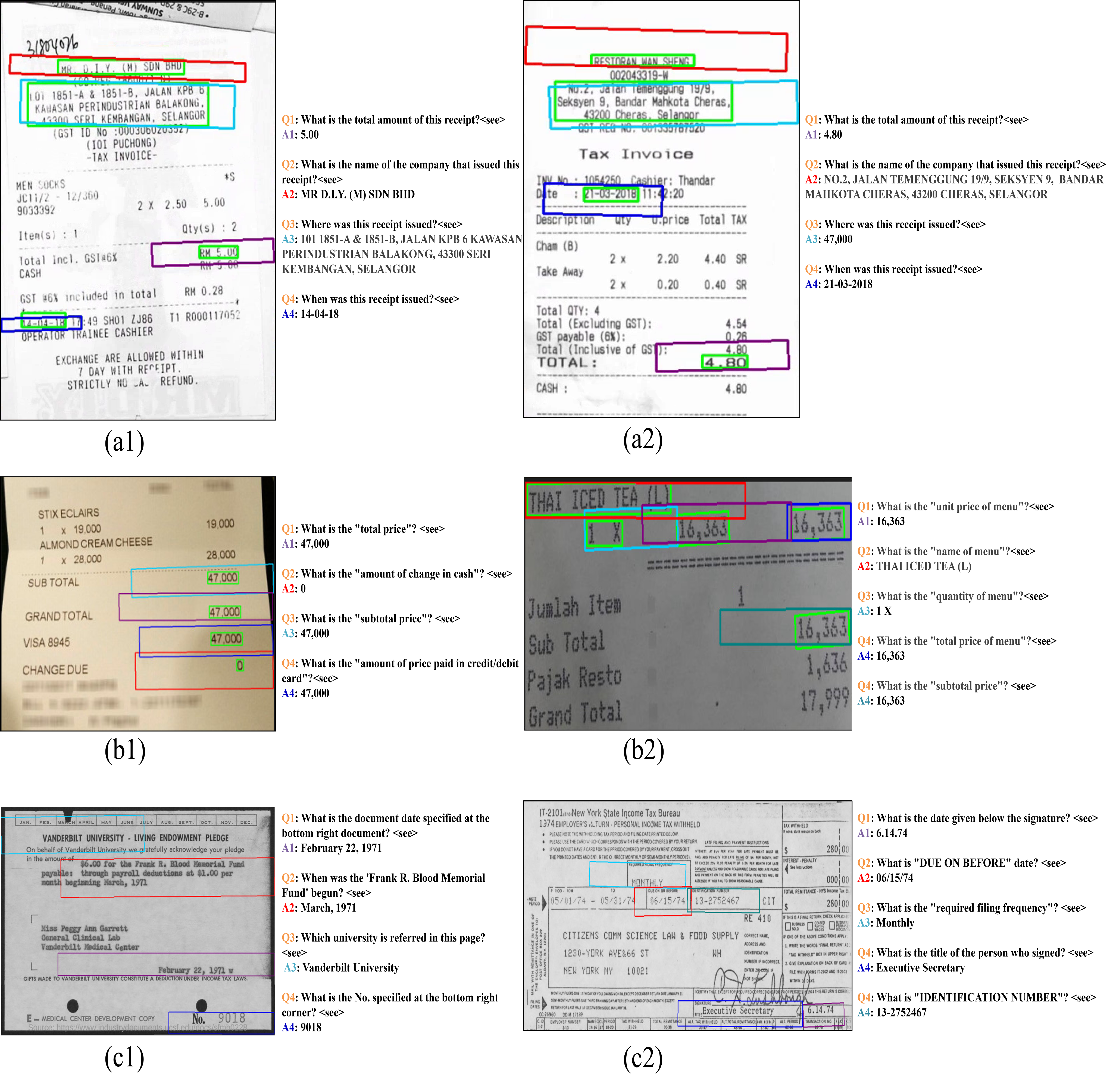}
	\caption{The results of acquiring text coordinates by Qwen2-VL-2B with $\mathit{see}$. a* refer to SROIE, b* refer to CORD, and c* refer to DocVQA. Different colors of the polygon boxes represent answers to different questions. }
	\label{fig:results_qwen}
\end{figure*}

As shown in Table~\ref{tab:MLLM-improvement}, our method yields consistent performance gains on both Qwen2-VL-2B and Qwen2-VL-7B, surpassing the improvements achieved by previous methods such as RIDGE~\cite{RIDGE}. These results demonstrate the strong generalizability of our approach. Examples of the prompt formats and the text coordinates predicted by Qwen2-VL-2B with $\mathit{see}$ are shown in Figure~\ref{fig:results_qwen}. 

\begin{table}[h]
	\centering
	\begin{tabular}{l|cccc}
		\toprule
		Model  & CORD & SROIE & DocVQA \\
		\midrule
		Qwen2-VL-7B~\cite{Qwen2-VL}  & 80.40 & 97.50 & 91.66 \\
		+ RIDGE~\cite{RIDGE}   & 85.53 & 97.74 & - \\
		+  $\mathit{see}$ (Ours)  & 85.83 & 97.92 & 91.97 \\
		\midrule
		Qwen2-VL-2B~\cite{Qwen2-VL}   & 76.76 & 92.64 & 84.80 \\
		+ $\mathit{see}$ (Ours)  & 79.59 & 93.79 & 86.80 \\
		\bottomrule
	\end{tabular}
	\caption{Performance improvements in zero-shot KIE for MLLMs. All results are evaluated using ANLS.}
	\label{tab:MLLM-improvement}
\end{table}

\subsection{Ablation Study}
To validate the effectiveness of each of our contributions, we build systems T1 through T4 based on STNet, which is built on the Donut architecture. We evaluate all systems on the SROIE dataset.
\begin{table}[t]
	\centering
	\begin{tabular}{cccccc}
		\toprule  
		System & TVG & $\mathit{See}$ & SS & F1 & Acc. \\  
		\midrule  
		T1 & \ding{56} & \ding{56} & \ding{56}& 84.7 & 94.1 \\
		T2 & \ding{52} & \ding{56} & \ding{56}& 86.2 & 96.3 \\
		T3 & \ding{52} & \ding{52} & \ding{56}& 87.8 & 97.1 \\
		T4 & \ding{52} & \ding{52} & \ding{52}& 88.3 & 97.4 \\
		\bottomrule  
	\end{tabular}
	\caption{Results of the evaluation for the STNet model on the SROIE dataset. ``TVG” denotes the use of the TVG dataset. $\mathit{See}$ indicates the inclusion of the \texttt{<see>} token, and ``SS” signifies the use of $\mathit{see}$ supervision on downstream SROIE. ``F1” and ``Acc.” correspond to the performance metrics on the SROIE test set.}
	\label{tab:ablation}
\end{table}

\subsubsection{Impact of See Loss Weight.}
During STNet training, the total loss is computed as a weighted combination of $\mathscr{L}_{\mathrm{lm}}$ and $\mathscr{L}_{\mathrm{see}}$, which differ significantly in scale. Determining the optimal weight $\lambda$ for the $\mathit{see}$ loss is crucial for balancing these losses. As shown in Table~\ref{tab:see_loss}, $\lambda = 0.001$ achieves the best performance.

\begin{table}[h]
	\centering
	\begin{tabular}{ccccc}
		\toprule $\lambda$ & $1e^{-1}$ & $1e^{-2}$& $1e^{-3}$& $1e^{-4}$ \\
		\midrule
		F1&86.4& 88.0& 88.3& 88.0 \\
		\bottomrule
	\end{tabular}
	\caption{Comparison of STNet's performance under different $\mathit{see}$ loss weights $\lambda$.}
	\label{tab:see_loss}
\end{table}

\subsubsection{Effectiveness of the TVG Dataset.}
The TVG dataset is designed to enhance the training efficacy of our models. To determine whether performance improvements stem solely from extended pre-training, we conduct two experimental setups, T1 and T2, as detailed in Table~\ref{tab:ablation}. The results show that T2 significantly outperforms T1, validating the effectiveness of the TVG dataset.

\subsubsection{Effectiveness of ``See then Tell''.} 
As shown in Figure~\ref{fig:results}, our STNet is capable of providing vision grounding for the answers, highlighting its ability to $\mathit{see}$. To evaluate whether this ability results in improved extraction accuracy, we compare the performance of T2 and T3, as shown in Table~\ref{tab:ablation}. The results indicate a marked improvement in T3 over T2, which doesn’t generate \texttt{<see>} embedded with text coordinates for physical location perception to guide the output of answer text. This validates the effectiveness of ``see then tell''.

\subsubsection{Robustness without Downstream See Supervision.}
\begin{figure}[t]
	\centering
	\includegraphics[width=\linewidth]{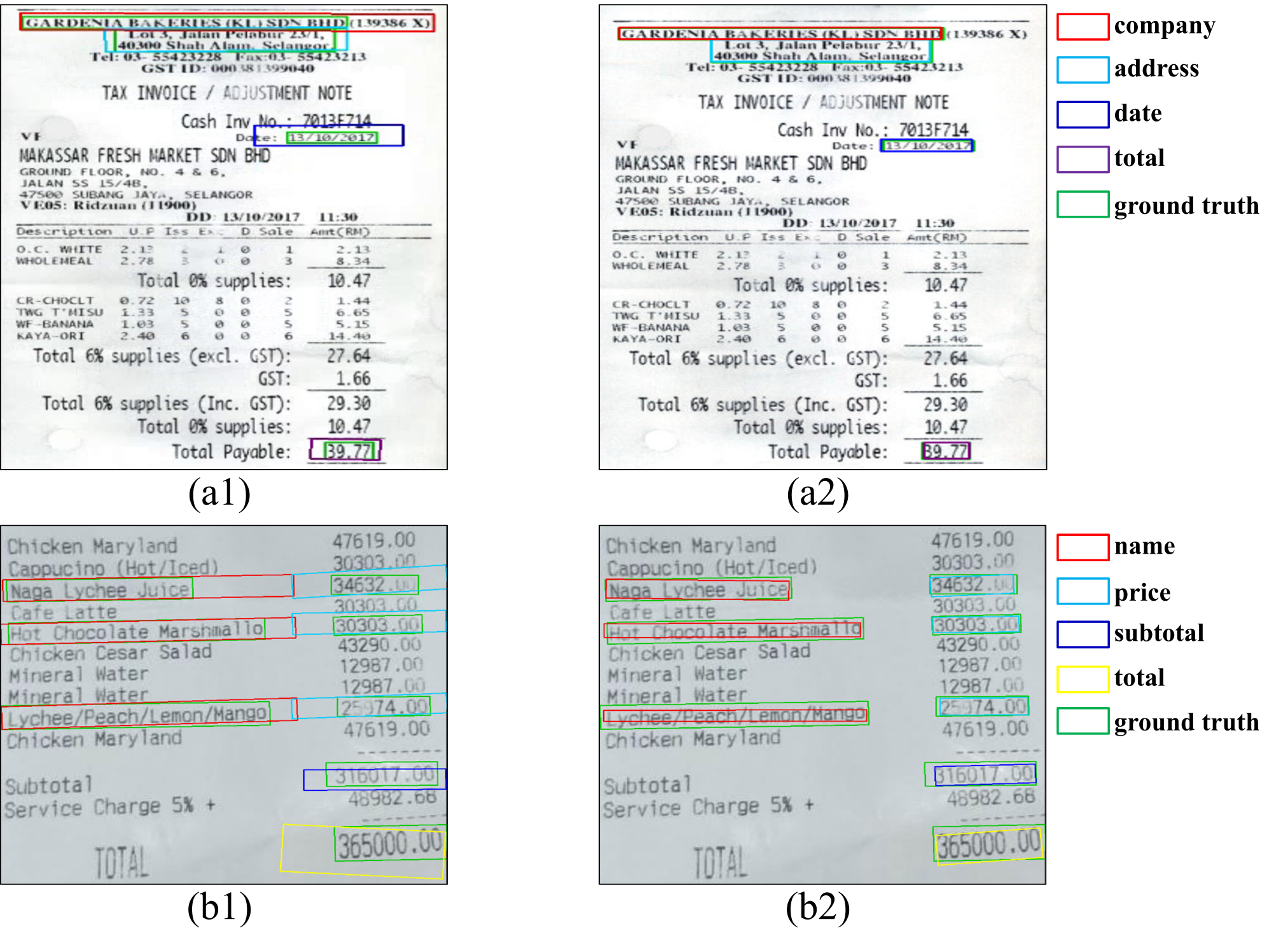}
	\caption{Comparison of text coordinates acquired by T3 and T4. a* refer to SROIE, b* refer to CORD. (a1) and (b1) refer to T3, (a2) and (b2) refer to T4. For clarity, some polygon boxes are omitted. }
	\label{fig:ablation}
\end{figure}

The calculation of $\mathit{see}$ loss requires answer location annotations, which are typically unavailable in many downstream document datasets such as DocVQA. To assess whether our model depends on such supervision during fine-tuning, we compare T3 (trained without downstream $\mathit{see}$ supervision) and T4 (trained with it). As shown in Table~\ref{tab:ablation}, T3 achieves performance very close to T4. Although Figure~\ref{fig:ablation} shows that T3’s predicted coordinates are less precise, they still reliably cover the correct answer regions. Furthermore, the IoU test on predicted boxes (Table~\ref{tab:iou}) confirms that T3 can still approximate the correct answer positions. These results demonstrate that STNet remains effective without relying on downstream $\mathit{see}$ supervision, highlighting its robustness in practical scenarios where location annotations are not available.

\begin{table}[h]
	\centering
	\begin{tabular}{cccc}
		\toprule  
		Threshold & $1e^{-3}$ & $1e^{-2}$ & $1e^{-1}$ \\  
		\midrule  
		T3 & 86.3 & 85.5 & 80.6  \\
		T4 & 97.6 & 97.6 & 97.0  \\
		\bottomrule  
	\end{tabular}
	\caption{Accuracy results for polygon box predictions. A prediction is considered correct if the IoU exceeds the defined threshold.}
	\label{tab:iou}
\end{table}

\section{Conclusion}
In this work, we introduce STNet, a novel end-to-end model that not only provides textual answers but also excels in offering vision grounding. STNet employs a ``see then tell" strategy, first outputting a special \texttt{<see>} token that encodes the answer's coordinates within the image as vision grounding to guide subsequent text generation. A dedicated physical decoder and a corresponding $\mathit{see}$ loss are designed to decode and supervise these coordinates. To effectively train \texttt{<see>}, we collect a number of table recognition datasets and develop a GPT-4-driven automated QA pair generation method, resulting in the TVG dataset, which comprises QA pairs with precise vision grounding. Experimental results on publicly available datasets such as CORD, SROIE, and DocVQA demonstrate that our method achieves state-of-the-art performance in Key Information Extraction. It generalizes well without access to downstream coordinate annotations during fine-tuning, showing robustness in practical scenarios where such annotations are unavailable. Furthermore, the proposed vision grounding mechanism can be seamlessly integrated into Multimodal Large Language Models, such as Qwen2-VL, leading to consistent improvements in zero-shot settings.

\bibliographystyle{elsarticle-num}
\bibliography{refs}


%
%
%

\end{document}